%% file: neurips_2020.tex
\documentclass{article}

\PassOptionsToPackage{numbers, compress}{natbib}

\usepackage[final]{neurips_2020}
\usepackage[utf8]{inputenc} 
\usepackage[T1]{fontenc}    
\usepackage{hyperref}       
\usepackage{url}            
\usepackage{booktabs}       
\usepackage{amsfonts}       
\usepackage{nicefrac}
\usepackage{algorithm}
\usepackage{algorithmic}
\usepackage{caption}
\usepackage{graphicx}
\usepackage{textcomp}
\usepackage{amsmath,amssymb}
\usepackage{comment}
\usepackage{pifont}

\newcommand{\xmark}{\ding{55}}

\newcommand{\repparams}{\theta}	
\newcommand{\taskparams}{W}

\newcommand{\loss}{\ell}

\newcommand{\task}{\mathcal{T}}

\newcommand{\Softmax}{\textit{Softmax}}

\usepackage{microtype}      

\usepackage{subfig}
\usepackage[export]{adjustbox}

\title{Few-Shot Unsupervised Continual Learning through Meta-Examples}
\author{Alessia Bertugli \thanks{Equal contribution.} \\
  University of Trento\\
  \texttt{alessia.bertugli@unitn.it} \\
  \And
  Stefano Vincenzi~\footnotemark[\value{footnote}]\\
  University of Modena and Reggio Emilia \\
  \texttt{stefano.vincenzi@unimore.it} \\
  \AND
  Simone Calderara \\
  University of Modena and Reggio Emilia \\
  \texttt{simone.calderara@unimore.it} \\
  \And
  Andrea Passerini \\
  University of Trento \\
  \texttt{andrea.passerini@unitn.it} \\
}

\begin{document}

\maketitle

\input{sections/0_abstract}

\input{sections/1_introduction}
\input{sections/2_tasks}
\input{sections/3_meta-attention}

\input{sections/4_experiments}
\input{sections/5_related}
\input{sections/6_discussion}

\bibliographystyle{splncs.bst}
\bibliography{references.bib}
\input{sections/7_supplementary}

\end{document}

%% file: sections/0_abstract.tex
\begin{abstract}
In real-world applications, data do not reflect the ones commonly used for neural networks training, since they are usually few, unlabeled and can be available as a stream. Hence many existing deep learning solutions suffer from a limited range of applications, in particular in the case of online streaming data that evolve over time. To narrow this gap, in this work we introduce a novel and complex setting involving unsupervised meta-continual learning with unbalanced tasks. These tasks are built through a clustering procedure applied to a fitted embedding space. We exploit a meta-learning scheme that simultaneously alleviates catastrophic forgetting and favors the generalization to new tasks. Moreover, to encourage feature reuse during the meta-optimization, we exploit a single inner loop taking advantage of an aggregated representation achieved through the use of a self-attention mechanism. Experimental results on few-shot learning benchmarks show competitive performance even compared to the supervised case.
Additionally, we empirically observe that in an unsupervised scenario, the small tasks and the variability in the clusters pooling play a crucial role in the generalization capability of the network. Further, on complex datasets, the exploitation of more clusters than the true number of classes leads to higher results, even compared to the ones obtained with full supervision, suggesting that a predefined partitioning into classes can miss relevant structural information. The code is available at \url{https://github.com/alessiabertugli/FUSION-ME}
\end{abstract}

%% file: sections/1_introduction.tex
\section{Introduction}
\label{sec:introduction}
Continual learning has been widely studied in the last few years to solve the catastrophic forgetting problem that affects neural networks.
Several methods~\cite{ewc, gem,a-gem,mer, si,iCaRL} have been proposed to solve this problem involving a replay buffer, network expansion, selectively regularizing and distillation. Some works~\cite{reconciling,MetaCL,itaml,ilmbml,cmlwt,hsml} take advantage of the meta-learning abilities of generalization on different tasks and rapid learning on new ones to deal with continual learning problems.
Few works on unsupervised meta-learning~\cite{cactus,umtra,uflst} and unsupervised continual learning~\cite{curl} have been recently proposed, but the first ones deal with independent and identically distributed data, while the second one assumes the availability of a huge dataset.
\begin{figure*}[!t]
    \centering
    \includegraphics[width=\linewidth]{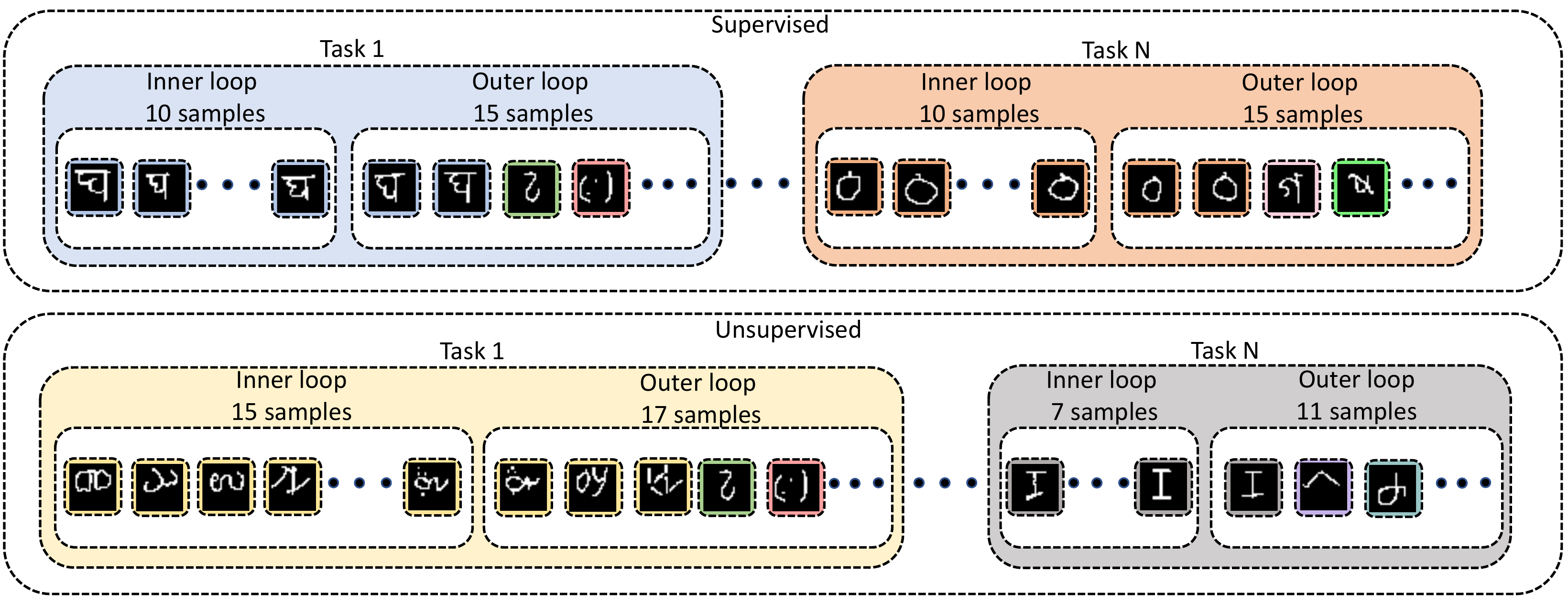}
    \caption{Supervised vs unsupervised tasks flow. In the supervised version, tasks are perfectly balanced and contain a fixed number of elements for inner loop (10 samples) and outer loop (15 samples, 5 from the current cluster and 10 randomly sampled from other clusters). In the unsupervised model, tasks are unbalanced and contain $2/3$ of cluster data for the inner loop and $1/3$ for the outer loop in addition to a fixed number of random samples.}
    \label{fig:tasks}
\end{figure*}
Moreover, the majority of continual learning and meta-learning works assume that data are perfectly balanced or equally distributed among classes.
We propose a new, more realistic setting, namely FUSION (Few-shot UnSupervIsed cONtinual learning), dealing with unlabeled and unbalanced tasks in a meta-continual learning fashion and a novel method MEML (Meta-Example Meta-Learning), that is able to face this complex scenario.
In the task construction phase, rather than directly exploiting high dimensional raw data, an embedding learning network is used to learn a fitted embedding space to facilitate clustering. Precisely, the k-means algorithm is applied to build tasks composed of unbalanced data, each one with the assigned pseudo-label. Our meta-learning model relies on a double-loop procedure that receives data in an online incremental learning fashion.
The classification layers are learned through a single inner loop update, adopting an attentive mechanism that extracts the most relevant features -meta example- of the current unbalanced task; this considerably reduces the training time and memory usage. In the outer loop, to avoid forgetting and improve generalization, we train all model layers exploiting, as input, an ensemble between data of the same class of the stream and data randomly sampled from the overall trajectory (see Figure~\ref{fig:tasks}).
We test our model and setup on Omniglot~\cite{omniglot} and Mini-ImageNet~\cite{imagenet}, achieving favorable results compared to baseline approaches.
We show the importance of performing the single inner loop update on the meta-example with respect to both updating over a random sample and updating over multiple samples of the same task. We empirically verify that with tasks generated in an unsupervised manner, the need for balanced data is not crucial compared to the variability in the data and the exploitation of small clusters.

%% file: sections/2_tasks.tex
\section{Few-Shot Unsupervised Continual Learning}
We propose a novel setting that deals with unsupervised meta-continual learning and study the effect of the unbalanced tasks derived by an unconstrained clustering approach.

As done in~\cite{cactus}, the task construction phase exploits the k-means algorithm over suitable embeddings obtained through an unsupervised pre-training. This simple but effective method assigns the same pseudo-label to all data points belonging to the same cluster. 
The first step employs two different models: Deep Cluster~\cite{deepcluster} for Mini-ImageNet, and ACAI~\cite{acai} for Omniglot. Both these methods consist of unsupervised training and produce an embedding vector set $Z = Z_{0}, Z_{1},..., Z_{N}$, where N is the number of data points in the training set. 
ACAI is based on an autoencoder while Deep Cluster on a deep feature extraction phase followed by k-means clustering. They outline some of the most promising approaches to deal with unlabeled, high dimensional data to obtain and discover meaningful latent features.
Applying k-means over these embeddings leads to unbalanced clusters, which determine unbalanced tasks. This is in contrast with typical meta-learning and continual learning problems, where data are perfectly balanced. To recover a balanced setting, in \cite{cactus}, the authors set a threshold on the cluster dimension, discarding extra samples and smaller clusters. A recent alternative~\cite{selflabelling} forces the network to balance clusters, but this imposes a partitioning of the embedding space that contrasts with the extracted features.
We believe that these approaches are sub-optimal as they alter the data distribution. In an unsupervised setting, where data points are grouped based on the similarity of their features, variability is an essential factor. By keeping also the small tasks, our model generalizes better and reaches higher accuracy at meta-test time.
In a data imbalanced setting, the obtained meta-representation is more influenced by large clusters. Since the latter may contain more generic features than the smaller ones, the model is able to generalize better by mostly learning from them. Despite this, the small clusters may contain important information for different classes presented during evaluation.
To corroborate this claim, we investigate balancing techniques, both at data-level, such as data augmentation and at model-level, such as balancing parameters into the loss term.
Once the tasks are built, they are sampled one at a time for the meta-continual train. The training process happens in a class incremental way, where a task correspond to one cluster. During this phase, the network have to learn a good representation that will be able to generalize to unseen tasks, while avoiding forgetting. The meta-train phase relies completely on pseudo-labels.
At meta-continual test time, novel and unseen tasks are presented to the network. The representation learned during meta-train remains fixed, while only the prediction layers are fine-tuned, testing on few data of novel classes.

%% file: sections/3_meta-attention.tex
\section{Meta-Example Meta-Learning}
\label{sec:mte-att}
Our network is composed of a Feature Extraction Network (FEN) and a CLassification Network (CLN), both updated during the meta-training phase through a meta-learning procedure based on the construction of a meta-example.
MAML and all its variants rely on a two-loop mechanism that allows learning new tasks from a few steps of gradient descent. Recent investigations on this algorithm explain that the real reason for MAML's success resides in feature reuse instead of rapid learning~\cite{anil}, proving that learning meaningful representations is a crucial factor.
Based on this assumption, we focus on the generalization ability of the feature extraction layers.
We remove the need for several inner loops, maintaining a single inner loop update through an attentive procedure that considerably reduces the training time and computational resources needed for training the model and increases the global performance.
At each time-step, as pointed out in Figure~\ref{fig:tasks}, a task $\mathcal{T}_{i} = (\mathcal S_{cluster}, \mathcal S_{query})$ is randomly sampled from tasks distribution $p(\mathcal{T})$. $\mathcal S_{cluster}$ contains elements of the same cluster and is defined as $\mathcal S_{cluster} = \{(X_{k}, Y_{k})\}_{k=0}^K, \; \text{with} \; Y_{0} = ... = Y_{K}$, where $Y_{0} = ... = Y_{k}$ is the cluster pseudo-label.
Instead, $\mathcal S_{query}$ contains a variable number of elements belonging to the current cluster and a fixed number of elements randomly sampled from all other clusters, and is defined as $\mathcal S_{query} = \{(X_{q}, Y_{q})\}_{q=0}^Q$.
All the elements belonging to $\mathcal S_{cluster}$  are processed by the frozen FEN, parameterized by $\theta$, computing the feature vectors $R_{0}, R_{1}, ..., R_{K}$ in parallel for all task elements as $R_{0:K} = f_\theta(X_{0:K})$.
The obtained embeddings are refined with an attention function parameterized by $\rho$ computes the attention coefficients $\alpha$ from the features vectors: 
\begin{equation}
    \label{eq:alpha}
    \alpha_{0:K} = \Softmax[f_\rho(R_{0:K})].
\end{equation}
Then, the final aggregated representation learning vector $ME$, called \emph{meta-example}, captures the most salient features, and is computed as follows:
\begin{equation}
    \label{eq:sum}
    ME = \displaystyle \sum_{k=0}^K [R_{k}*\alpha_{k}].
\end{equation}
The single inner loop is performed on this meta-example, which adds up the weighted-features contribution of each element of the current cluster. Then, the cross-entropy loss $\loss$ between the predicted label and the pseudo-label is computed and both the classification network parameters $W$ and the attention parameters $\rho$ ($\psi = \{\taskparams_i, \rho\}$) are updated as follows:
\begin{equation}
    \label{eq:inner_loop}
    \psi \leftarrow \psi-\alpha
		\nabla_{\psi} \loss_i(f_{\psi}(\text{\textit{ME}}), Y_{0}),
\end{equation}
where $\alpha$ is the inner loop learning rate.
Finally, to update the whole network parameters $\phi =\{\repparams, \taskparams_{i}, \rho\}$, and to ensure generalization across tasks, the outer loop loss is computed $S_{query}$. The outer loop parameters are thus updated as follows:
\begin{equation}
    \label{eq:outer_loop}
    \phi \leftarrow \phi - \beta \nabla_\phi \loss_i (f_{\phi}( X_{0:Q}), Y_{0:Q}),
\end{equation}
where $\beta$ is the outer loop learning rate.

%% file: sections/4_experiments.tex
\section{Experiments}
\label{sec:experiments}
\subsection{Balanced \textit{vs.} Unbalanced Tasks}
To justify the use of unbalanced tasks and show that allowing unbalanced clusters is more beneficial than enforcing fewer balanced ones, we present in Table~\ref{tab:omni} some comparisons achieved on the Omniglot dataset.
First of all, we introduce a baseline in which the number of clusters is set to the true number of classes, removing from the task distribution the ones containing less than $N$ elements and sampling $N$ elements from the bigger ones (OML). We thus obtain a perfectly balanced training set at the cost of less variety within the clusters; however, this leads to poor performance as small clusters are never represented. Setting a smaller number of clusters than the number of true classes gives the same results (OML balanced 500). This test shows that cluster variety is more important than balancing for generalization.
To verify if maintaining variety and balancing data can lead to better performance, we try two balancing strategies: augmentation, at data-level, and balancing parameter, at model-level. For the first one, we keep all clusters, sampling $N$ elements from the bigger and using data augmentation for the smaller to reach $N$ elements (OML augmentation). 
At model-level, we multiply the loss term by a balancing parameter, to weight the update for each task based on cluster length (OML balancing param).
These two tests, especially the latter one, result in lower performance with respect to the unbalanced setting, suggesting that the only thing that matters is cluster variety. We can also presume that bigger clusters may contain the most meaningful and general features, so unbalancing does not negatively affect the training of our unsupervised meta-continual learning model.
Finally, as we want to confirm that this intuition is valid in a more general unsupervised meta-learning model, we perform the balanced/unbalanced experiments also on CACTUs~\cite{cactus}. The results are shown in Table~\ref{tab:unbalmaml} (Top) and attest that the model trained on unbalance data outperforms the balanced one, further proving the importance of task variance to better generalize to new classes at meta-test time. We report the results training the algorithms on $20$ ways for generality purposes and $5$ shots and $15$ shots, in order to have enough data points per class to create the imbalance.

\subsection{Meta-example Single Update \textit{vs.} Multiple Updates}
In Table~\ref{tab:omni}, we show that the model trained with the attention-based method consistently outperforms all the other baselines. The single update gives the worst performance, but not really far from the multiple updates one, confirming the idea that the strength of generalization relies on the feature reuse. Also, the mean test has performance comparable with the multiple and single update ones, proving the effectiveness of the attention mechanism to determine a suitable and general embedding vector for the CLN.
Training time and resources consumption is considerably reduced with our model based on a single update on the generated meta-example (see Supplementary Material).
We also test our technique in a standard meta-learning setting. We compare our meta-example based algorithm MEML to MAML~\cite{maml} on Omniglot dataset in Table~\ref{tab:unbalmaml} (bottom), consistently outperforming it. We report the results training on $20$ ways and $1$ and $5$ shots. In particular, the $5$ shots test highlights the effectiveness of our aggregation method.

\begin{table}[t]
\begin{minipage}{.48\textwidth}
\caption{Meta-test results on Omniglot.}
\label{tab:omni} 
\resizebox{\textwidth}{!}{
    \begin{tabular}{l|cccccc}
     \toprule\noalign{\smallskip}
     Algorithm/Classes   & 10 & 50 & 75 & 100 & 150 & 200 \\
     \noalign{\smallskip} \toprule
     Oracle OML~\cite{oml} & 88.4 & 74.0 & 69.8  & 57.4 & 51.6  & 47.9  \\ 
     Oracle MEML & \textbf{92.3}  & \textbf{78.2}  & \textbf{72.7}  & \textbf{60.9}  & \textbf{51.8}  & \textbf{51.4} \\ \toprule
     OML balanced 500 & 67.8 & 27.6 & 29.4 & 24.5 & 18.7 & 15.8  \\
     OML balancing param & 59.4 & 27.2 & 24.3 & 18.4 & 15.5 & 11.8 \\
     OML augmentation & 72.2 & 35.1 & 32.5 & 27.5 & 21.8 & 17.3 \\ \toprule
     OML & 74.6 & 32.5 & 30.6 & 25.8 & 19.9 & 16.1 \\
     OML single update & 67.5 & 32.0 & 30.2  & 24.3  & 18.4  & 15.3  \\
     MEML mean & 60.6  & 31.2 & 25.8  & 21.3 & 17.0 & 13.7  \\
     MEML (Ours) & \textbf{84.6 } & 37.3  & 37.5 & 30.9  & 25.4  & 20.7 \\
     MEML RS (Ours) & \underline{81.6} & \textbf{56.4} & \textbf{54.0 } & \textbf{44.6 } & \textbf{34.1 } & \textbf{27.4} \\
     \bottomrule
    \end{tabular}
    }
\end{minipage}
\begin{minipage}{.47\textwidth}
\caption{Meta-test results on Mini-ImageNet.}
\label{tab:min}
\resizebox{\textwidth}{!}{
     \begin{tabular}{l|ccccc}
     \toprule\noalign{\smallskip}
     Algorithm/Classes   & 2 & 4 & 6 & 8 & 10 \\
     \noalign{\smallskip} \toprule
     Oracle OML~\cite{oml}  & 50.0 & 31.9 & 27.0  & 16.7  & 13.9  \\
     Oracle MEML  & \textbf{66.0} & \textbf{33.0} & \textbf{28.0} & \textbf{29.1} & \textbf{21.1} \\ \toprule
     OML & 49.3 & 41.0  & 19.2  & 18.2 & 12.0 \\    
     MEML 64 & 58.0  & 41.2 & \textbf{40.0}  &  27.3  & 18.8 \\ 
     MEML 128 & 56.0 & 41.7   & 21.6 & 16.2 & 11.4 \\
     MEML 256  & \textbf{70.0}  & \textbf{48.4} & \underline{36.0}  & \textbf{34.0} & \textbf{21.6}  \\
     MEML 512 & 54.7  & 36.4  & 26.2 & 14.1  & 21.4 \\
     MEML 64 RS  & 54.0 & 39.0 & 31.2 & 27.3  & 16.4  \\     
     \bottomrule
     \end{tabular}
     }
\end{minipage}
\end{table}

\begin{table}[t]
\centering
    \caption{Balanced \textit{vs.} unbalanced CACTUs-MAML (top) and MEML, with our meta-example update, compared to basic MAML (bottom) on Omniglot dataset.}
     \begin{tabular}{l|cccc}
      \toprule\noalign{\smallskip}
     Algorithm/Ways, Shots   & 5,1 & 5,5 & 20,1 & 20,5 \\
     \noalign{\smallskip} \toprule
     CACTUs-MAML Balanced 20,5 & 60.50 & 84.00 & 40.50 & 67.62 \\
     CACTUs-MAML Unbalanced 20,5 &	\textbf{62.50} & \textbf{85.50} & \textbf{42.62} & \textbf{71.87} \\ \toprule
     CACTUs-MAML Balanced 20,15 & 67.00	& 86.00	& 32.50 & 64.62 \\
     CACTUs-MAML Unbalanced 20,15 & \textbf{72.00}	& \textbf{89.00}	& \textbf{40.00}	& \textbf{66.25} \\ \toprule \toprule
     MAML 20,1  & 78.00 &	97.50 & 77.62 &	92.87 \\ 
     MEML 20,1 (Ours) & \textbf{97.50} &	\textbf{99.97} &	\textbf{88.13} &	\textbf{99.37} \\ \toprule
     MAML 20,5  & 88.00	 &	99.50 &	74.62 &	92.75 \\ 
     MEML 20,5 (Ours) & \textbf{95.00} & \textbf{99.95} &	\textbf{85.63} &	\textbf{96.25} \\ \toprule
     \end{tabular}
     \label{tab:unbalmaml}
\end{table}

\subsection{MEML \textit{vs.} Oracles}
To see how the performance of MEML is far from those achievable with the real labels, we also report for all datasets the accuracy reached in a supervised setting (\emph{oracles}) on both Omniglot (see Table~\ref{tab:omni}) and Mini-ImageNet (see Table~\ref{tab:min}).
We define Oracle OML the supervised model present in \cite{oml}, and Oracle MEML the supervised model updated with our meta-example strategy. Oracle MEML outperforms Oracle OML on Omniglot and Mini-ImageNet, suggesting that the meta-examples strategy is beneficial even in a fully supervised case.
MEML reaches higher performance compared to the other OML baselines but lower on Omniglot compared to the Oracle OML. On Mini-ImageNet, our model trained with $256$ clusters outperforms both oracles.
To further improve the performance avoiding forgetting at meta-test time, we add a rehearsal strategy based on reservoir sampling on the CLN (MEML RS). This generally results in superior performance on Omniglot. On Mini-ImageNet the performance with and without rehearsal are similar, due to the low number of test classes in the dataset that alleviates catastrophic forgetting. 

\subsection{Number of Clusters}
In an unsupervised setting, the number of original classes could be unknown. Consequently, it is important to assess the performance of our model by varying the number of clusters at meta-train time. With a coarse-grain clustering, a low number of clusters are formed and distant embeddings can be assigned to the same pseudo-label, grouping classes that can be rather different. On the other hand, with a fine-grain clustering, a high number of clusters with low variance are generated. Both cases lead to poor performance at meta-test time.
We test on Omniglot (see Table~\ref{tab:omni}), setting the number of clusters to: the true number of classes (OML); a lower number of clusters (OML balanced 500), resulting in more than 20 samples each.
Since the Omniglot dataset comprehends 20 samples per class, in the first case it results in unbalanced tasks, while in the second we sample 20 elements from the bigger clusters.
The performance of the 1100 clusters test is consistently higher than that obtained with the 500 clusters test, confirming that variability is more important than balancing.
On Mini-ImageNet, we test our method in Table~\ref{tab:min} with 64, 128, 256, and 512 clusters (MEML number of clusters). Since Mini-ImageNet contains 600 examples per class, after clustering we sample examples between 10 and 30, proportionally to the cluster dimension. We obtain the best results with 256 clusters and the meta-example approach, outperforming not only the other unsupervised experiment but also the supervised oracle.
We observe that using 512 clusters degrades performance with respect to the 256 case, suggesting that tasks constructed over an embedding space with too specific features fail to generalize. Using a lower number of clusters, such as 64 or 128, also achieves worse performance. This time, the embedding space is likely aggregating distant features, leading to a complex meta-continual training, whose pseudo-classes are not clearly separated. 

%% file: sections/5_related.tex
\section{Related Work}
\label{sec:related}
\subsection{Supervised and Unsupervised Continual Learning}
Continual learning is one of the most challenging problems arising from neural networks that are heavily affected by catastrophic forgetting. The proposed methods can be divided into three main categories. \emph{Architectural strategies}, are based on specific architectures designed to mitigate catastrophic forgetting~\cite{pnn,prog_comp}. \emph{Regularization strategies} are based on putting regularization terms into the loss function, promoting selective consolidation of important past weights~\cite{ewc,si}. Finally \emph{rehearsal strategies} focus on retaining part of past information and periodically replaying it to the model to strengthen connections for memories, involving meta-learning~\cite{mer,Spigler2019MetalearntPS}, combination of rehearsal and regularization strategies~\cite{gem,a-gem}, knowledge distillation~\cite{BornAN,lwf,hou2018lifelong,lee2019overcoming}, generative replay~\cite{genrep,Silver_2020_CVPR_Workshops,Liu2020GenerativeFR} and channel gating~\cite{cvpr_davide}.
Only a few recent works have studied the problem of unlabeled data, which mainly involves representation learning. CURL~\cite{curl} proposes an unsupervised model built on a representation learning network. This latter learn a mixture of Gaussian encoding task variations, then integrates a generative memory replay buffer as a strategy to overcome forgetting.

\subsection{Supervised and Unsupervised Meta-Learning}
Meta-learning, or learning to learn, aims to improve the neural networks ability to rapidly learn new tasks with few training samples. The majority of meta-learning approaches proposed in literature are based on Model-Agnostic Meta-Learning (MAML)~\cite{maml,mlig,reptile,leo}. Through the learning of a profitable parameter initialization with a double loop procedure, MAML limits the number of stochastic gradient descent steps required to learn new tasks, speeding up the adaptation process performed at meta-test time.
Although MAML is suitable for many learning settings, few works investigate the unsupervised meta-learning problem. CACTUs~\cite{cactus} proposes a new unsupervised meta-learning method relying on clustering feature embeddings through the k-means algorithm and then builds tasks upon the predicted classes.
UMTRA~\cite{umtra} is a further method of unsupervised meta-learning based on a random sampling and data augmentation strategy to build meta-learning tasks, achieving comparable results with respect to CACTUs.
UFLST~\cite{uflst} proposes an unsupervised few-shot learning method based on self-supervised training, alternating between progressive clustering and update of the representations.

\subsection{Meta-Learning for Continual Learning}
Meta-learning has extensively been merged with continual learning for different purposes. We can highlight the existence of two strands of literature~\cite{osaka}: \emph{meta-continual learning} with the aim of incremental task learning and \emph{continual-meta learning} with the aim of fast remembering.
Continual-meta learning approaches mainly focus on making meta-learning algorithms online, with the aim to rapidly remember meta-test tasks~\cite{oml_finn,reconciling,cmlwt}.
More relevant to our work are meta-continual learning algorithms~\cite{MetaCL,oml,l2cl,hsml,ilmbml,itaml,FewShotCL}, which use meta-learning rules to ``learn how not to forget". OML~\cite{oml} and its variant ANML~\cite{l2cl} favor sparse representations by employing a trajectory-input update in the inner loop and a random-input update in the outer one. The algorithm jointly trains a representation learning network (RLN) and a prediction learning network (PLN) during the meta-training phase. Then, at meta-test time, the RLN layers are frozen and only the PLN is updated. ANML replaces the RLN network with a neuro-modulatory network that acts as a gating mechanism on the PLN activations following the idea of conditional computation.

%% file: sections/6_discussion.tex
\section{Discussion}
\label{sec:discussion}
In this work, we tackle a novel problem concerning few-shot unsupervised continual learning. We propose a simple but effective model based on the construction of unbalanced tasks and meta-examples.
Our model is motivated by the power of \emph{representation learning}, which relies on few and raw data with no need for human supervision. With an unconstrained clustering approach, we find that no balancing technique is necessary for an unsupervised scenario that needs to generalize to new tasks. In fact, the most robust and general features are gained though task variety; even if favoring larger clusters leads to more general features, smaller ones should not be discarded as they can be representative of less common tasks.
This means that there is no need for complex representation learning algorithm that try to balance clusters elements. A future achievement is to deeply investigate this insight by observing the variability of the embeddings in the feature space.
A further improvement consists in the introduction of FiLM layers~\cite{perez2017film} into the FEN to change data representation at meta-test time and the introduction of an OoD detector to face with Out-of-Distribution tasks.
The performances of our model with meta-examples suggest that a single inner update can increase performances if the most relevant features for the task are selected. To this end, a more refined technique, relying on hierarchical aggregation techniques, can be considered.

%% file: sections/7_supplementary.tex
\linewidth\hsize \vskip 0.625in minus 0.125in 
\onecolumn{\LARGE\bf \centering Supplementary Material \\ \par}
\vspace{1cm}
\appendix
\section{Test on SlimageNet64}
We also make some preliminary attempts on SlimageNet64~\cite{Antoniou2020DefiningBF}, a novel and difficult benchmark for few-shot continual learning. We make the embeddings using DeepCluster~\cite{deepcluster} and we report the obtained results in Table~\ref{tab:slimagenet}. We find that our update method based on meta-example overcomes the baselines on both supervised and unsupervised approaches. MEML with 1600 clusters reaches better performances then the 800 clusters cases, meaning that a more refined partition of the embedding space is more beneficial then a rigid partitioning into classes.

\begin{table}[H]
\centering
     \caption{Meta-test test results on SlimageNet64 dataset.}
     \begin{tabular}{l|cccccc}
      \toprule\noalign{\smallskip}
     Algorithm/Classes   & 5 & 10 & 20 & 30 & 40 & 50 \\
     \noalign{\smallskip} \toprule
     Oracle OML & 24.0 \textpm 2.5. & 14.3 \textpm 3.1 & 15.9 \textpm 8.3 & 5.0 \textpm 1.1 & \textbf{9.4 \textpm 1.6} & 2.0 \textpm 0.4\\
     Oracle MEML & \textbf{31.2 \textpm 1.6} & \textbf{23.8 \textpm 1.7} & \textbf{25.3 \textpm 2.9} & \textbf{5.5 \textpm 1.4} & \underline{7.2 \textpm 1.3} & \textbf{5.7 \textpm 0.3} \\ \toprule
    OML & 22.4 \textpm 1.9 & 16.1 \textpm 1.9 & 23.3 \textpm 5.0 & 3.8 \textpm 0.6 & \underline{8.9 \textpm 0.0} & 2.1 \textpm 1.2 \\
    MEML 800 & 22.4 \textpm 3.2 & 13.9 \textpm 0.0 & 25.8 \textpm 0.0 & \textbf{4.7 \textpm 0.3} & \textbf{9.0 \textpm 0.1} & 2.2 \textpm 0.4 \\
    MEML 1600  & \textbf{27.2 \textpm 1.6} & \textbf{16.1 \textpm 1.1} & \textbf{26.4 \textpm 0.6} & \underline{4.2 \textpm 0.6} & 8.1 \textpm 1.0 & \textbf{2.3 \textpm 0.7} \\
    \toprule
    \end{tabular}
    \label{tab:slimagenet}
\end{table}

\section{Out-of-Distribution Tasks}
Since our model is unsupervised, FEN training is only based on feature embeddings, with no class-dependent bias. This way, our model could be general enough for OoD tasks, where the training tasks belong to a different data distribution (i.e. a different dataset) with respect to the test tasks.
To investigate this conjecture, we test our model on the Cifar100 and Cub datasets. Results in Table~\ref{tab:ds_cifar_cub} show that, by training on Mini-ImageNet and testing on Cifar100 (top half) or training on Omniglot and testing on Cub (bottom half), the unsupervised approach generally outperforms the supervised one. In the latter case, MEML also outperforms the supervised oracle trained on Cub, which is incapable of learning a meaningful representation in our particular setting.

\begin{table}[H]
\centering
 \caption{Meta-test test results with Out-of-Distribution tasks on Cifar100 and Cub datasets.}
     \begin{tabular}{l|ccccc}
      \toprule\noalign{\smallskip}
     \textbf{Cifar100}/Classes & 2 & 4 & 6 & 8 & 10 \\
     \noalign{\smallskip} \toprule
     Oracle OML Cifar100 & 66.0 & 45.0 & 34.0 & 30.0 & 29.5  \\ \toprule
     Oracle MEML Mini-ImageNet & 58.0 & 33.0 & \textbf{35.3} & 25.7 & \textbf{24.9} \\
     MEML Mini-ImageNet & \textbf{66.0} & \textbf{35.0} & \underline{28.7} & \textbf{34.3} &  \underline{22.2} \\ \hline \toprule 
     \textbf{Cub}/Classes & 2 & 10 & 20 & 30 & 40 \\ \toprule
     Oracle OML Cub & 50.0 & 13.9 & 25.8 & 4.5 & 8.9  \\  \toprule
     Oracle MEML Omniglot & 44.0 & 49.1 & 32.7 & \textbf{27.0} & 25.1 \\
     MEML Omniglot & \textbf{66.0} & \textbf{53.3} & \textbf{28.3} & \underline{26.2} & \textbf{25.6}  \\
     \toprule
     \end{tabular}
     \label{tab:ds_cifar_cub}
\end{table}
\section{Rehearsal at Meta-Train Time}
Rehearsal strategy can be useful at meta-test time. In particular, when the CLN is adapted to new tasks in an incremental fashion, its weights can be overridden favoring the last tasks at the expense of the first ones. The beneficial effect of rehearsal at meta-test time can be noticed when the number of test tasks is high. In fact, reservoir sampling is generally helpful on Omniglot, that is tested on 200 classes, while it does not give the same benefit on Mini-ImageNet, where it reaches similar or a little lower performance.
We want to verify if rehearsal can be beneficial also at meta-train time, replacing the query set $\mathcal S_{query}$ with a coreset built with reservoir sampling $\mathcal S_{coreset}$. This way, instead of sampling from random clusters, a buffer of previously seen data is stored in a buffer of fixed dimension. We try three different memory size 200, 500, 1000, obtaining, as expected, increasing results as the size increases.
In Table~\ref{tab:omni_rs} we report accuracy results on Omniglot adding rehearsal only at meta test time, and adding it at both meta-train and meta-test time with OML and MEML. We report only the results obtained with a buffer size 500 to avoid redundancies. As it can be noted, with OML, using a coreset instead of a query set at meta-train time increase the performance with respect to the case of query set usage, meaning that the representation suffers from catastrophic forgetting and the use of random data (acting only for generality purposes and not contrasting forgetting) are not enough to learn a good representation.
On the contrary, with MEML, the use of a rehearsal strategy at meta-train time get worse performance.
We hypothesize that this behavior is due to the different number of inner loop update between the two models. In fact, OML, making several inner loop updates on data belonging to the same cluster, brings the CLN weights nearest the current cluster, suffering the effect of forgetting more then MEML that makes a single inner loop on the meta-example. These results prove that, at meta-train time, MEML needs only the \emph{generalization} ability given by $\mathcal S_{query}$, while OML needs also the \emph{remembering} ability given by $\mathcal S_{coreset}$.

\begin{table}[!t]
 \caption{Meta-test test results on Omniglot dataset with rehearsal only during meta-test and both at meta-train and meta-test.}
 \centering
     \begin{tabular}{l|cccccc}
     \toprule\noalign{\smallskip}  
     Algorithm/Classes   & 10 & 50 & 75 & 100 & 150 & 200 \\
     \noalign{\smallskip} \toprule 
     OML RS only test & 67.9 & 55.1 & 46.2 & 37.0 & 29.6 & \textbf{25.6} \\
     OML RS both train/test & \textbf{75.9} & \textbf{56.8} & \textbf{51.2} & \textbf{39.7} & \textbf{30.5} & \underline{25.0} \\ \toprule
     MEML RS only test & \textbf{81.6} & \textbf{56.4} & \textbf{54.0} & \textbf{44.6} & \textbf{34.1}  & \textbf{27.4} \\
     MEML RS both train/test & 74.7 & 47.0 & 48.4 & 38.3 & 28.9 & 24.2\\
     \toprule
     \end{tabular}
     \label{tab:omni_rs}
\end{table}

\section{Details on Balancing Techniques}
To verify the effect of unbalanced tasks during meta-training, we apply two techniques to balance tasks, one at data-level, \emph{data augmentation} and the other at model-level, \emph{loss balancing}. We briefly explain how these methods are implemented.
\subsection{Data Augmentation}
We apply data augmentation on the Omniglot dataset to observe if balancing the clusters could lead to superior performance. We notice that the results reached applying data augmentation are comparable with the one obtained with unbalanced tasks. 
Practically, we sample 20 elements from the clusters bigger than 20, while we exploit augmentation on the cluster with less than 20 elements. Till reaching 20 samples for tasks, we pick each time a random image between the ones in the cluster employing a random combination of various augmentation techniques, such as horizontal flip, vertical flip, affine transformations, random crop, and color jitter. In detail, about the random crop, we select a random portion included between $75\%$, $80\%$, $85\%$, or $90\%$ of the entire image. Regarding the color jitter, we use brightness, contrast, saturation, and hue factor (the first three denote a factor including between $0.8$ and $1.2$, the hue instead one including between $-0.02$ and $0.02$) to adjust the image.  
\subsection{Loss Balancing}
Our model applies clustering on all training data before starting to learn the meta-representation. This way, we can find the maximum $C_{max}$ and minimum number $C_{min}$ of elements per cluster obtained by k-means algorithm. Then, for each cluster, we find its number of elements $C_{current}$ and compute the balanced vector $\Gamma$ as follow.
\begin{equation}
    \label{eq:bal_vec}
    \Gamma = \frac{C_{max} - C_{min}}{C_{current} - C_{min} + \epsilon},
\end{equation}
where $\epsilon$ is used to avoid division by zero. Finally $\Gamma$ is normalized as follow.
\begin{equation}
    \label{eq:norm_bal_vec}
    \Gamma_{norm} = \frac{\Gamma - \Gamma_{min}}{\Gamma_{max} - \Gamma_{min}}.
\end{equation}
For each sampled task ($\text{taskId}$), the corresponding balancing parameter is selected and multiplied by the cross-entropy loss $CE$ during meta-optimization as reported in below.
\begin{equation}
    \label{eq:loss_bal}
    L =  \Gamma_{norm}[\text{taskId}] \cdot CE(\text{logits}, Y),
\end{equation}
where $\text{logits}$ indicate the output of the model.
\section{Comparison with SeLa Embeddings}
We try a recent embedding learning method based on self-labeling, SeLa~\cite{selflabelling}, that forces a balanced separation between clusters. In Table~\ref{tab:min_sela}, we report the results obtained training our model with SeLa embeddings on Mini-ImageNet.
The main idea, taking up what was done in DeepCluster~\cite{deepcluster}, is to join clustering and representation learning, combining cross-entropy minimization with a clustering algorithm like K-means. This approach could lead to degenerate solutions such as all data points mapped to the same cluster.
The authors of SeLa tried to solve this issue by adding the constraint that the labels must induce equipartition of the data, which they observe maximizes the information between data indices and labels. This new criterion extends standard cross-entropy minimization to an optimal transport problem, which is harder to optimize, exploiting traditional algorithms that scale badly when facing larger datasets. To solve this problem a fast version of the Sinkhorn-Knoop algorithm is applied. 

In detail, given a dataset of  $N$ data points $I_1,\dots,I_N$ with corresponding labels $\mathbf{y}_1,\dots,\mathbf{y}_N \in \{1,\dots,K\}$, drawn from a space of $K$ possible labels, and a deep neural network $\mathbf{x} = \Phi(I)$ mapping $I$ to feature vectors $\mathbf{x}\in\mathbb{R}^D$; the learning objective is defined as:
\begin{equation}
\begin{split}
  \min_{p,q} & \ E(p,q) \\
  \quad\text{subject to}\
  \forall \mathbf{y}:&q(\mathbf{y}|\mathbf{x}_i) \in \{0,1\} ~\text{and} \\
  ~\sum_{i=1}^N q(\mathbf{y}|&\mathbf{x}_i) = \frac{N}{K}.
\end{split}
\label{eq:sela_obj}
\end{equation}
$E(p,q)$ is defined as the average cross-entropy loss, while the constraints mean that the $N$ data points are split uniformly among the K classes and that each $\mathbf{x}_i$ is assigned to exactly one label.
The objective in Equation~(\ref{eq:sela_obj}) is solved as an instance of the optimal transport problem, for further details refer to the paper.
\begin{table}[!t]
 \caption{Meta-test test results on Mini-ImageNet dataset with Sela embedding.}
 \centering
     \begin{tabular}{l|ccccc}
      \toprule\noalign{\smallskip}
     Algorithm/Classes   & 2 & 4 & 6 & 8 & 10 \\
     \noalign{\smallskip} \toprule
     OML & 50.0 & 25.0 & \textbf{18.0} & 31.3 & 15.0 \\    
     MEML 64 & 50.0  & \textbf{35.0} & 17.3  &  31.3  & 17.0 \\ 
     MEML 256  & \textbf{64.0} & 31.0 & 17.3 & \textbf{32.5} & \textbf{18.3}  \\
     \toprule
     \end{tabular}
     \label{tab:min_sela}
\end{table}
DeepCluster adopts particular implementation choices to avoid degenerate solutions, but contrary to SeLa it does not force the clusters to contain the same number of samples.
We empirically observe that in our setting an unconstrained approach leads to better results.

\section{Time and Computational Analysis}
In Table~\ref{tab:time_memory}, we compare training time and computational resources usage between OML and MEML on Omniglot and Mini-ImageNet. Both datasets confirm that MEML, adopting a single inner update, trains considerably faster and uses approximately one-third of the GPU resources with respect to OML. This latter performs an update for each sample included in $\mathcal S_{cluster}$, keeping a computational graph of the model in memory for each update. This leads to slower training time, especially when the required number of epochs is high, such as for Mini-ImageNet. Even though with this kind of datasets we do not require particular GPU resources, this test shows the strength of our model in an eventually future scenario exploiting large image, deeper network, and more cluster samples.
\begin{table}[!t]
 \caption{Training time and GPU usage of MEML \textit{vs.} OML on Omniglot and Mini-ImageNet.}
 \centering
     \begin{tabular}{lccc}
     \toprule\noalign{\smallskip}
     Model & Dataset & Training time & GPU usage \\
     \hline\addlinespace[0.05cm]
     OML & Omniglot & 1h 32m  & 2.239 Gb \\
     MEML & Omniglot & 47m & 0.743 Gb \\
     OML & Mini-ImageNet & 7h 44m & 3.111 Gb \\
     MEML & Mini-ImageNet & 3h 58m & 1.147 Gb \\
     \noalign{\smallskip} \toprule
     \end{tabular}
     \label{tab:time_memory}
\end{table}

\section{Learning in the Jungle}
To the best of our knowledge, a few-shot unsupervised continual learning setting has never been studied before in the literature. However, some works propose ``learning in the jungle" problems, that involve a mixture of non-trivial settings. In Table~\ref{tab:comparison} we compare some novel methods to our OML, highlighting the features of each one. Our model is the only one that presents such complex setting involving few-shot learning, continual learning, unlabelled and unbalanced tasks and proposes experiments that show the model ability to learn from OoD data. Note that this analysis is not intended to be a complete analysis of all the methods of continual learning and few-shot learning, but only of those methods that have been placed in a different setting from the one that is commonly used in these two fields or that are related to them.

\begin{table*}[!t]
 \caption{Features comparison between our MEML and several works recently proposed in the literature involving continual learning and few-shot learning into the wild.}
 \centering
     \begin{tabular}{ccccc|l}
     \toprule\noalign{\smallskip}
      Few-shot & Unsupervised & Continual & Imbalance & OoD & Algorithm \\
      \xmark & \xmark & \checkmark & \xmark & \xmark & iCARL~\cite{iCaRL} \\
      \xmark & \checkmark & \checkmark & \xmark & \xmark & CURL~\cite{curl} \\
      \checkmark & \checkmark & \xmark & \xmark & \xmark & CACTUs~\cite{cactus} \\
      \checkmark & \checkmark & \xmark & \xmark & \xmark & UMTRA~\cite{umtra} \\
      \checkmark & \checkmark & \xmark & \xmark & \xmark & UFLST~\cite{uflst} \\
      \checkmark & \xmark & \xmark & \checkmark & \checkmark & L2B~\cite{l2b}\\
      \xmark & \checkmark & \checkmark & \xmark & \checkmark & GD~\cite{lee2019overcoming}\\
      \checkmark & \xmark & \checkmark & \xmark & \xmark & OML~\cite{oml}\\
      \checkmark & \xmark & \checkmark & \xmark & \xmark & ANML~\cite{l2cl}\\
      \checkmark & \xmark & \checkmark & \xmark & \checkmark & Continual-MAML~\cite{osaka}\\
      \checkmark & \xmark & \checkmark & \xmark & \xmark & iTAML~\cite{itaml}\\
      \checkmark & \checkmark & \checkmark & \checkmark & \checkmark & MEML (Ours) \\
     \noalign{\smallskip} \toprule
     \end{tabular}
     \label{tab:comparison}
\end{table*}

\section{FiLM Layers for OoD Tasks}
To further improve the results testing on OoD tasks, we introduce FiLM~\cite{perez2017film} layers within the OML architecture (the supervised baseline). In a \emph{FiLMed} neural network some conditional input is used to conditioned FiLM layers, to influence the final prediction by this input. The FiLM generator map this information into FiLM parameters, applying  feature-wise affine transformation (in particular scaling and shifting) element-wise (features map-wise for CNN). If $\mathbf{x}$ is the input of a FiLM layer, $\mathbf{z}$ a conditional input, and $\gamma$ and $\beta$ are $\mathbf{z}$-dependent scaling and shifting vectors, the FiLM transformation is reported below.
\begin{equation}
    \label{eq:film}
    \text{FiLM} = \gamma(\mathbf{z}) \mathbf{x} \odot \beta(\mathbf{z})
\end{equation}

We apply this concept to OML, conditioning the prediction to task-specific features. We add two FiLM layers as linear layers after each of the last two convolutional layers of the FEN. These layers have adaptable parameters, updating in both the inner and the outer loop. In detail, recovering what was already done in ~\cite{cavia}, we introduce a $100$-dimensional context parameter vector producing, through the linear layer, $512$ filters. These filters are used to apply an affine transformation on the output of the convolutional layer.
Context parameters are reset to zero before each new task, while FiLMs are trained to be general for all tasks and never reset during meta-train. 

At meta-test time, we update the FiLM layers (during the meta-test train phase) and we reset the context parameters after each new task. This way, the context parameters are specific and dependent on each task while the FiLM layers can adapt themselves to the new unseen classes, in order to shift the frozen representation according to the context.
This way, if a task changes, the model could be able to shift the representation reaching better generalization capabilities. The advantage is more pronounced facing with OoD tasks since their distribution is much different with respect to the meta-train one. We report some preliminary results obtained applying FiLM layers on the OML~\cite{oml} model, trained on Omniglot and tested on both Omniglot (see Table~\ref{tab:film_omni}) and Cifar100 (see Table~\ref{tab:film_cifar}). We find that OML with FiLM layers outperforms or at least equals on both dataset.

The results are promising, but we believe that much better performance could be achieved training context parameters and FiLM layers separately or introducing some tricks to train them together.  
\begin{table}[t]
 \caption{Meta-test test results on Omniglot dataset with FiLM layers applied on Oracle OML.}
 \centering
     \begin{tabular}{l|cccccc}
     \toprule\noalign{\smallskip}
     Algorithm/Classes   & 10 & 50 & 75 & 100 & 150 & 200 \\
     \noalign{\smallskip} \toprule 
     Oracle OML & 88.4 & 74.0 & 69.8  & 57.4 & 51.6  & 47.9  \\
     OML FiLM & \textbf{91.1} & \textbf{79.5} & \textbf{80.6} & \textbf{68.6} & \textbf{64.0} & \textbf{52.7} \\
     \toprule
     \end{tabular}
     \label{tab:film_omni}
\end{table}
\begin{table}[t]
 \caption{Meta-test test results on Cifar100 dataset with FiLM layers applied on Oracle OML trained on Omniglot.}
 \centering
     \begin{tabular}{l|ccccc}
      \toprule\noalign{\smallskip}
     \textbf{Cifar100}/Classes & 2 & 4 & 6 & 8 & 10 \\
     \noalign{\smallskip} \toprule
     OML Omniglot & \textbf{50.0} & \textbf{25.0} & 15.3 & 22.8 & 9.4  \\ \toprule
     OML FiLM Omniglot & \textbf{50.0} & \textbf{25.0} & \textbf{16.7} & \textbf{31.3} & \textbf{13.9} \\
     \toprule
     \end{tabular}
     \label{tab:film_cifar}
\end{table}

\section{The Effect of Self-Attention}
\begin{figure}
 \resizebox{\columnwidth}{!}{
     \centering
     \subfloat{\includegraphics[width=8cm, valign=c]{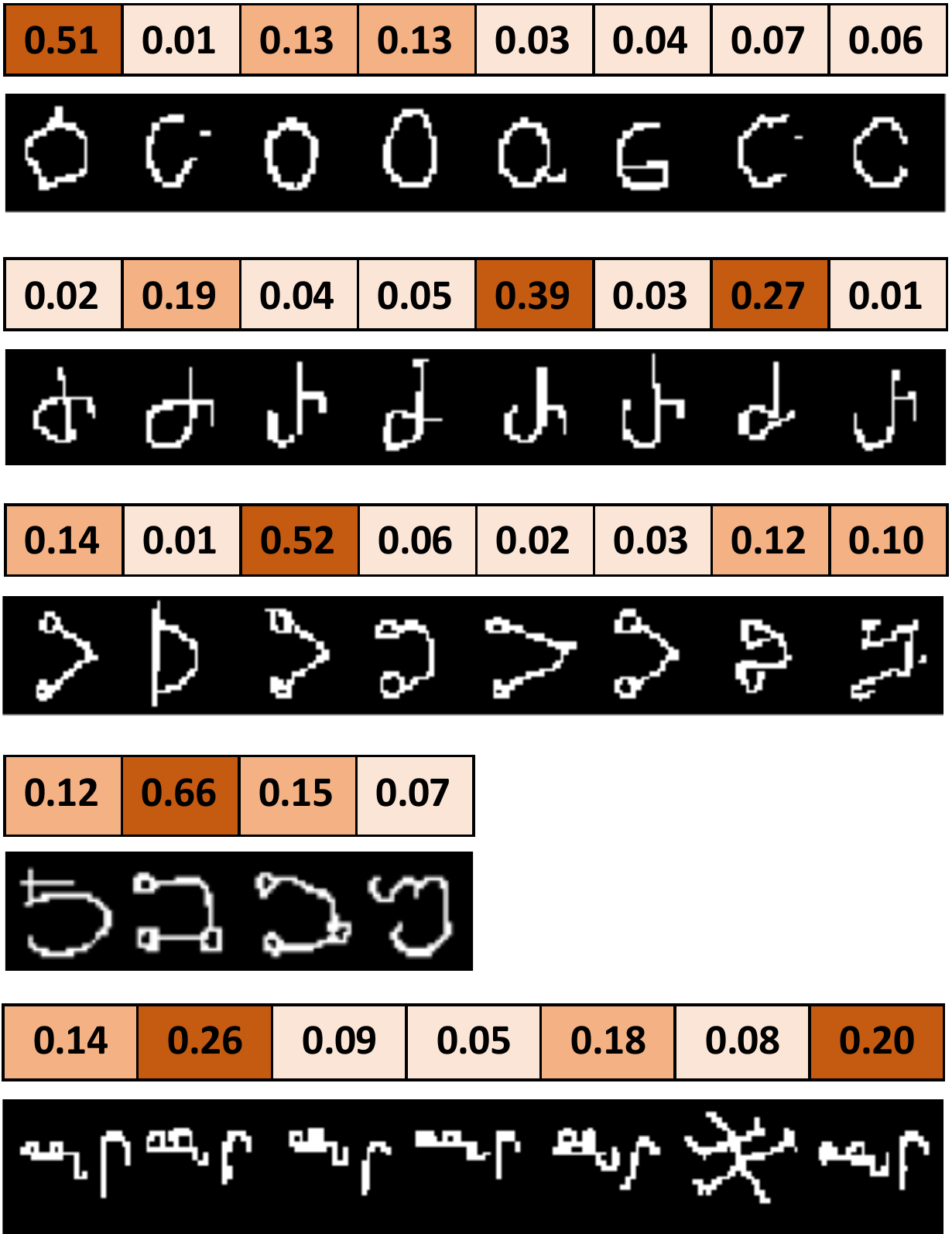}} \hspace{0.3cm}
     \subfloat{\includegraphics[width=10cm, valign=c]{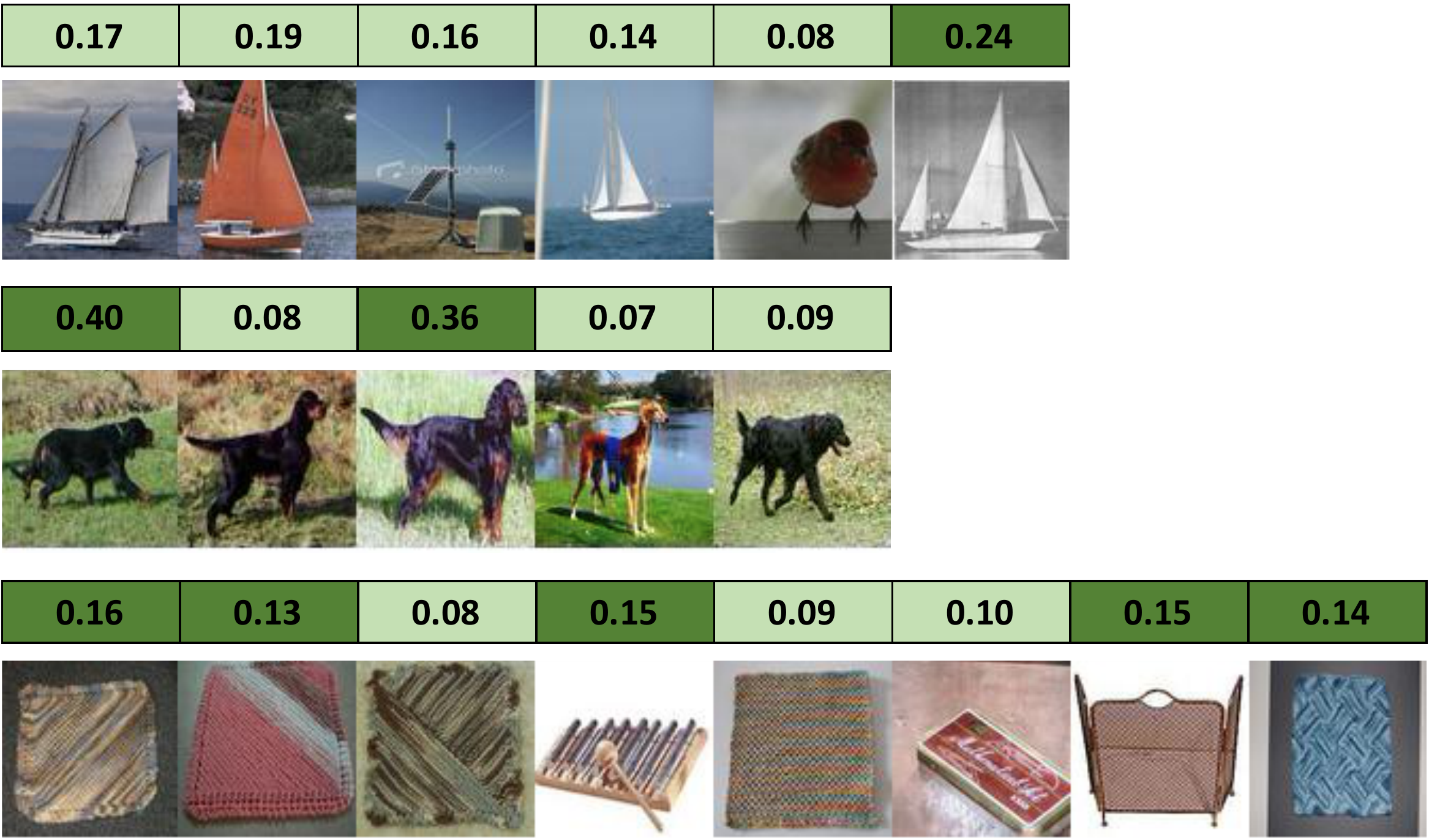}}}
     \caption{Samples of clusters (one for each row) generated on Omniglot (left) and Mini-ImageNet (right). Self-attention coefficients are reported associated to each image.}
     \label{fig:effect_att}
 \end{figure}

Here we want to empirically view how our self-attention mechanism acts on cluster images. We report some examples of clusters and the respectively self-attention coefficients that MEML associates to each image. In Figure~\ref{fig:effect_att} some samples obtained during MEML training are reported, on Mini-ImageNet and Omniglot respectively. The darker colors indicate the values of the highest attention coefficient, while the lighter colors indicate the lower ones. In the majority of cases, our mechanism rewards the most representative examples of the cluster, meaning the ones that globally contain most of the features present in the other examples as well. A further improvement could be to identify the outliers (the samples more distant from the others at features-level) of a cluster and discard them before the self-attention mechanism is applied. This way, only the features of the correctly grouped samples can be employed to build the meta-example.

\section{Datasets}
To evaluate our model, we employ two standard datasets typically used to validate few-shot learning methods: Omniglot and Mini-ImageNet. In addition, we try our model on a new and challenging few-shot continual learning benchmark, SlimageNet64.
The Omniglot dataset contains 1623 characters from 50 different alphabets with 20 greyscale image samples per class. We use the same splits as~\cite{cactus}, using 1100 characters for meta-training, 100 for meta-validation, and 423 for meta-testing.
The Mini-ImageNet dataset consists of 100 classes of realistic RGB images with 600 examples per class. As done in~\cite{meta-opt,cactus}, we use 64 classes for meta-training, 16 for meta-validation and 20 for meta-test. The SlimageNet64 dataset contains 1000 classes with 200 RGB images per class taken from the down-scaled version of ILSVRC-2012, ImageNet64x64. 800 classes are used for meta-train and the remaining 200 for meta-test purposes. Finally, we use the Cifar100~\cite{cifar100} and Cub~\cite{cub} datasets to prove our model performance on Out-of-Distribution tasks.

\section{Implementation Details}
\begin{figure}[H]
    \centering
    \includegraphics[width=10cm]{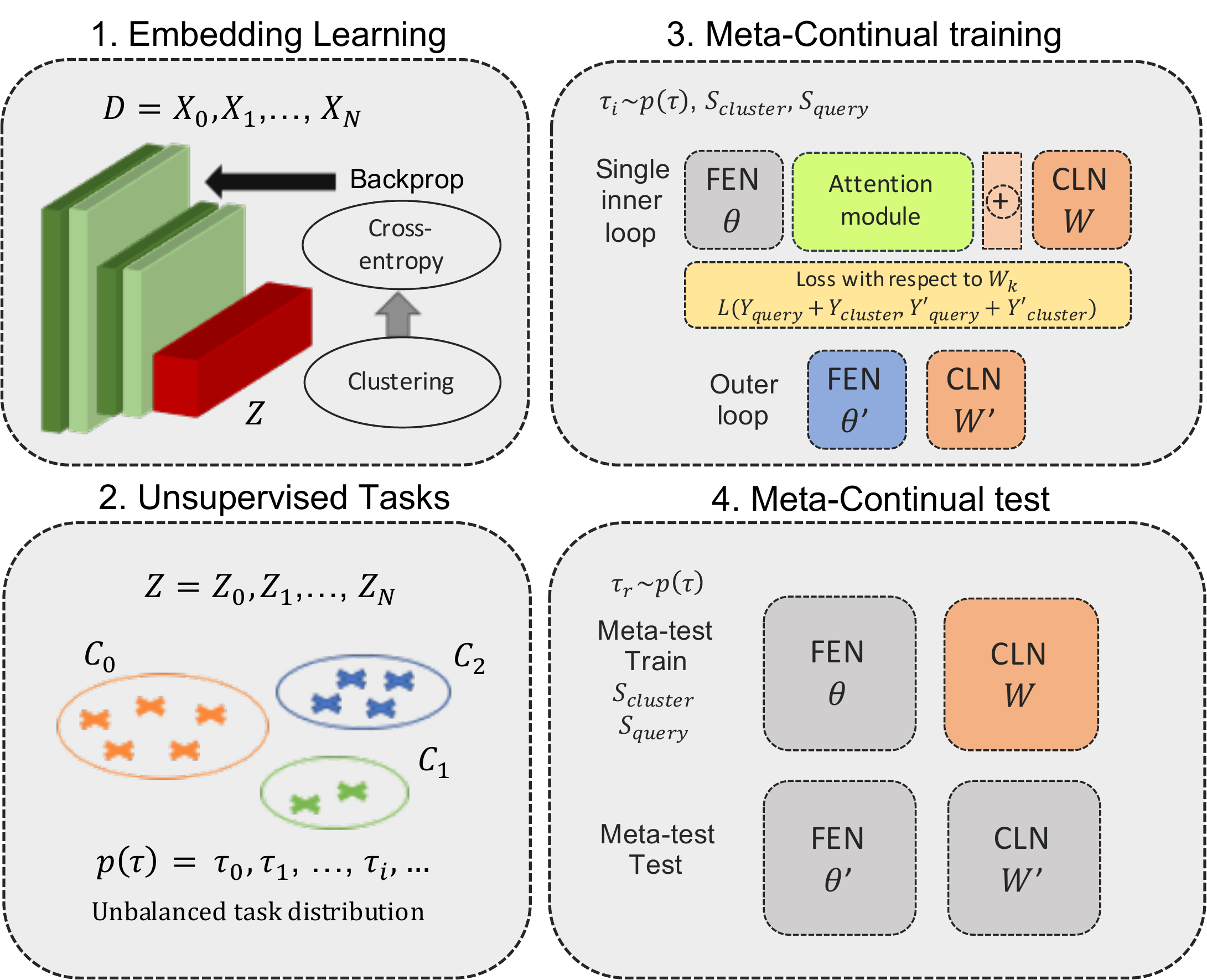}
    \caption{Scheme of FUSION. The model is composed of 4 phases: embedding learning network phase, unsupervised task construction phase, meta-continual training phase and meta-continual test phase.}
    \label{fig:model}
\end{figure}

The FEN is composed of $6$ convolutional layers followed by ReLU activations, $3\times3$ kernel (for Omniglot, the last one is a $1\times1$ kernel) followed by $2$ linear layers interleaved by a \textit{ReLU} activation. The attention mechanism is implemented with two additional linear layers interleaved by a \textit{Tanh} function and followed by a \textit{Softmax} and a sum to compute attention coefficients and aggregate features. For Omniglot, we train the model for $40000$ steps while for Mini-ImageNet and SlimageNet64 for $200000$, with meta-batch size equals to $1$. The outer loop learning rate is set to $1e^{-4}$ while the inner loop learning rate is set to $0.01$, with Adam optimizer. We report the algorithm of MEML meta-training in Algorithm~\ref{alg:fusion-me} and an illustration of the four phases in Figure~\ref{fig:model}.

\begin{algorithm}[!t]
	\centering
	\caption{MEML algorithm on FUSION setting}
	\label{alg:fusion-me}
	\begin{algorithmic}[1]
		\REQUIRE : $D = X_{0}, X_{1},..., X_{N}$: unlabeled training set
		\REQUIRE $\alpha$, $\beta$: inner loop and outer loop learning rates
		\STATE Run embedding learning on $D$ producing $Z_{0:N}$ from $X_{0:N}$
		\STATE Run $k$-means on $Z_{0:N}$ generating a distribution of unbalanced tasks $p(\task)$ from clusters
		\STATE Randomly initialize $\repparams$ and $\taskparams$
		\WHILE{not done}
		\STATE Sample a task $\task_i \sim p(\task) = (\mathcal S_{cluster}, \mathcal S_{query})$
		\STATE Randomly initialize $\taskparams_i$
		\STATE $\mathcal S_{cluster} = \{(X_{k}, Y_{k})\}_{k=0}^K$, with $Y_{0} = ... = Y_{K}$
		\STATE $\mathcal S_{query} = \{(X_{q}, Y_{q})\}_{q=0}^Q$
		\STATE $R_{0:K} = f_{\theta}(X_{0:K})$
		\STATE $\alpha_{0:K} = \Softmax[f_{\rho}(R_{0:K})]$
		\STATE $\text{\textit{ME}} = \displaystyle \sum_{k=0}^K [R_{k}*\alpha_{k}]$
		\STATE $\psi, \phi = \{\taskparams_i, \rho\}, \{\repparams, \taskparams_i, \rho\}$
        \STATE  $\psi \leftarrow \psi-\alpha
		\nabla_{\psi}  \loss_i(f_{\psi}(\text{\textit{ME}}), Y_{0})$ 
		\STATE $\phi \leftarrow \phi - \beta \nabla_\phi \loss_i (f_{\phi}( X_{0:Q}), Y_{0:Q})$ 
		\ENDWHILE
	\end{algorithmic}
\end{algorithm}
